\documentclass[runningheads]{llncs}
\usepackage{graphicx}

\usepackage{pifont}
\newcommand{\cmark}{\ding{51}}
\newcommand{\xmark}{\ding{55}}

\usepackage{tikz}
\usepackage{comment}
\usepackage{amsmath,amssymb}
\usepackage{color}

\usepackage[accsupp]{axessibility}  

\usepackage{booktabs}
\usepackage{multirow,multicol}
\usepackage{adjustbox}
\usepackage{cite}
\usepackage[inline]{enumitem}
\usepackage{wrapfig}
\usepackage{xspace}

\usepackage{xspace}
\usepackage{tabularx}

\makeatletter
\DeclareRobustCommand\onedot{\futurelet\@let@token\@onedot}
\def\@onedot{\ifx\@let@token.\else.\null\fi\xspace}

\def\eg{\emph{e.g}\onedot}

\def\etal{\emph{et al}\onedot}
\def\dataset{EgoProceL\xspace}
\def\datasetsp{EgoProceL }
\def\meth{CnC\xspace}
\def\methsp{CnC }
\def\tv{\widetilde{v}}

\newcommand{\myfirstpara}[1]{\noindent \textbf{#1:}}
\newcommand{\mypara}[1]{\myfirstpara{#1}}

\makeatother

\usepackage[pagebackref,breaklinks,colorlinks]{hyperref}
\usepackage[capitalize]{cleveref}
\crefname{section}{Sec.}{Secs.}
\Crefname{section}{Section}{Sections}
\Crefname{table}{Table}{Tables}
\crefname{table}{Tab.}{Tabs.}

\begin{document}

\pagestyle{headings}
\mainmatter
\def\ECCVSubNumber{1886}

\title{My View is the Best View: \\ Procedure Learning from Egocentric Videos}

\titlerunning{Procedure Learning from Egocentric Videos}

\author{
Siddhant Bansal\inst{1} \and
Chetan Arora\inst{2} \and
C.V. Jawahar\inst{1}
}

\authorrunning{S. Bansal et al.}

\institute{Center for Visual Information Technology, IIIT, Hyderabad \and
Indian Institute of Technology, Delhi\\
\email{siddhant.bansal@research.iiit.ac.in}}


\maketitle
\begin{abstract}
    Procedure learning involves identifying the key-steps and determining their logical order to perform a task. Existing approaches commonly use third-person videos for learning the procedure, making the manipulated object small in appearance and often occluded by the actor, leading to significant errors. In contrast, we observe that videos obtained from first-person (egocentric) wearable cameras provide an unobstructed and clear view of the action. However, procedure learning from egocentric videos is challenging because (a) the camera view undergoes extreme changes due to the wearer's head motion, and (b) the presence of unrelated frames due to the unconstrained nature of the videos. Due to this, current state-of-the-art methods' assumptions that the actions occur at approximately the same time and are of the same duration, do not hold. Instead, we propose to use the signal provided by the temporal correspondences between key-steps across videos. To this end, we present a novel self-supervised Correspond and Cut (\meth) framework for procedure learning. \methsp identifies and utilizes the temporal correspondences between the key-steps across multiple videos to learn the procedure. Our experiments show that \methsp outperforms the state-of-the-art on the benchmark ProceL and CrossTask datasets by $5.2\%$ and $6.3\%$, respectively. Furthermore, for procedure learning using egocentric videos, we propose the \datasetsp dataset consisting of $62$ hours of videos captured by $130$ subjects performing $16$ tasks. The source code and the dataset are available on the project page \href{https://sid2697.github.io/egoprocel/}{https://sid2697.github.io/egoprocel/}.
\end{abstract}

\section{Introduction}
\begin{figure}[t]
    \centering
    \includegraphics[width=\textwidth]{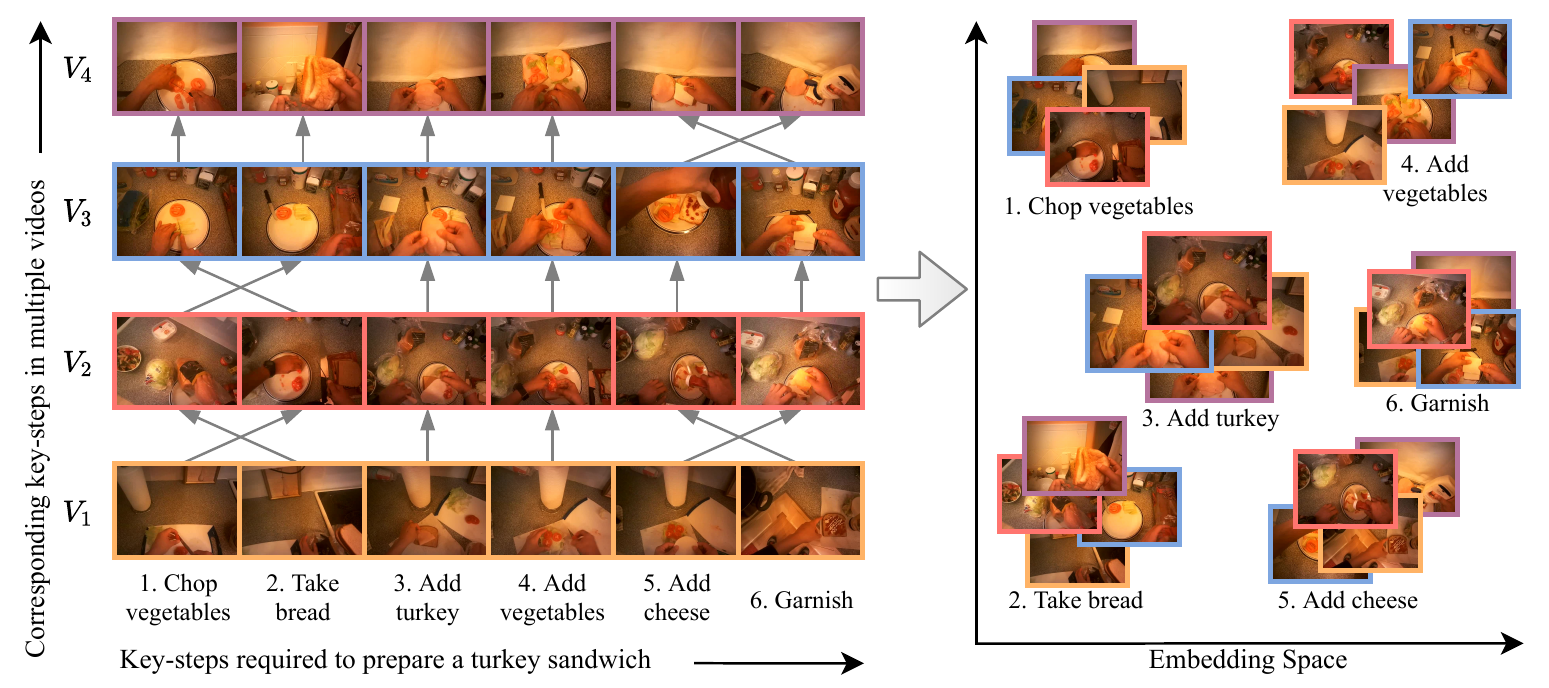}
    \caption{The left-hand side figure shows six key-steps required to prepare a turkey sandwich~\cite{egtea_gaze_p} across four egocentric videos. The arrows among the videos highlight the change in the ordering of corresponding key-steps. This work utilizes these correspondences and aims to learn an embedding space where the corresponding key-steps have similar embeddings (right-hand side figure). To this end, we propose \methsp which learns the embedding space and utilizes it to localize the key-steps and identify their ordering.}
    \label{fig:correspondences}
\end{figure}

Imagine showing an autonomous agent how to prepare a sandwich, and it learns the steps required for it!
Motivated by this vision, our work focuses on developing a framework that allows an agent to identify the steps required to perform a task and their order after observing multiple visual demonstrations by experts.
Given a set of instructional videos for the same task, procedure learning~\cite{multi-task-procl,joint_dynamic_summary,Shen_action_segmentation_2021_CVPR} broadly consists of two steps, (a) assigning all the frames to the $K$ key-steps (including the background), and (b) discovering the logical ordering of the key-steps required to perform the task.
Procedure learning differs from action segmentation as it aims to \textit{jointly} segment common key-steps (actions required to accomplish a task, as shown in \Cref{fig:correspondences}) across a given set of videos.
In contrast, action segmentation aims to identify actions (unrelated to their relevance to accomplishing a task) from a \textit{single} video.
Furthermore, procedure learning deals with additional or missing key-steps and background actions unrelated to the task and identifies an ordering of the key-steps.

Existing instructional videos datasets~\cite{Inria_dataset,joint_dynamic_summary,Breakfast,howto100m,COIN,YouCook2,CrossTask,Ji_2022_WACV} majorly consist of third-person videos.
Here, the camera is kept far from the expert, to avoid interference in the actual task. 
Due to this, the manipulated objects are typically small or sometimes invisible. 
Additionally, third-person videos can be captured from various positions, leading to wide variations in the camera viewpoints for the same task~\cite{CMU_Kitchens}.
Further, most datasets comprise videos scraped from the internet (YouTube)~\cite{joint_dynamic_summary,howto100m,COIN,CrossTask,Ji_2022_WACV}, which are noisy and have large irrelevant segments.
In contrast, egocentric cameras are typically harnessed to the subject's head and have a standardized location.
They provide a clearer view of the executed task, including the manipulated objects.
As a result, recent works have introduced datasets consisting of egocentric videos~\cite{Damen2018EPICKITCHENS,ego4d,tent,egtea_gaze_p,6248010,Sigurdsson2018CharadesEgoAL}, which have proven helpful for various tasks~\cite{Furnari2020RollingUnrollingLF,10.1007/978-3-030-01225-0_46,6751511,ng2019you2me,Singh_2016_CVPR}.

Motivated by the advantages of egocentric videos over third-person videos, we propose an egocentric videos dataset for procedure learning: \dataset.
\datasetsp consists of $62$ hours of egocentric videos of $16$ tasks ranging from making a salmon sandwich to assembling a Personal Computer (PC), thereby ensuring diversity of tasks and facilitating generalizable methods.
However, egocentric videos come with their own set of challenges.
For example, the camera view undergoes extreme movements due to the wearer's head motion, introducing frames unrelated to the activity and unavailability of the actor's pose~\cite{Singh_2016_CVPR}.

To overcome the challenges and learn the procedure from egocentric videos, we propose utilizing the signal provided by temporal correspondences across videos.
As shown in \Cref{fig:correspondences}, critical moments like putting a slice of turkey on the bread while preparing a turkey sandwich are present across all the videos.
To exploit the signal provided by such temporal correspondences, we propose a self-supervised, three-stage, Correspond and Cut (\meth) framework for procedure learning.
The first stage of the \meth uses the proposed self-supervised TC3I loss to learn an embedding space such that the same key-steps across the videos have similar embeddings (\Cref{fig:correspondences}).
The second stage consists of the proposed ProCut Module (PCM).
PCM performs clustering on the learned embeddings and assigns each frame to a key-step.
The final stage of \meth creates a key-step sequence for each video and infers relevant ordering to perform the task.

Current works mostly use frame-wise metrics to evaluate the models developed for procedure learning~\cite{multi-task-procl,joint_dynamic_summary,kukleva2019unsupervised,Shen_action_segmentation_2021_CVPR,VidalMata_2021_WACV}.
While these metrics evaluate the procedure reasonably well compared to simply calculating the accuracy, they do not suit datasets with significant class imbalance.
Furthermore, procedure learning datasets consist of significant background frames~\cite{CrossTask}.
Hence, a model assigning all the frames to the background might achieve high scores.
We propose to solve this problem by calculating the scores via the contribution of each key-step, leading to lower scores when models assign most of the frames to the background. 
Further, when comparing with the previous works, (a) we use \meth on standard third-person benchmark datasets~\cite{CrossTask,joint_dynamic_summary} and (b) employ existing metrics to evaluate.
We show that CnC outperforms the state-of-the-art techniques for procedure learning (\Cref{tab:table_third_person_results}).

\mypara{Contributions} The major contributions of our work are:
\begin{itemize}
    \item To facilitate procedure learning from egocentric videos, we create the \dataset dataset. The dataset consists of $62$ hours of videos captured by $130$ subjects performing $16$ tasks.
    \item We propose \meth, which utilizes the proposed TC3I loss and PCM to identify the key-steps and their ordering required to perform a task.
    \item We investigate the usefulness of egocentric videos over third-person videos for procedure learning. We observe an average improvement of $2.7\%$ in the F1-Score when using egocentric videos instead of third-person videos.
    \item The \dataset dataset and the code written for this work are released on \href{http://cvit.iiit.ac.in/research/projects/cvit-projects/egoprocel}{http://cvit.iiit.ac.in/research/projects/cvit-projects/egoprocel} (mirror link).
\end{itemize}

\section{Related Works}
We aim to perform procedure learning in a self-supervised fashion, unlike previous works~\cite{procedure_completion_BMVC_2020,Sener_2019_ICCV,YouCook2}, which assume the availability of mapping between video frames and key-steps.
Also, different from weakly supervised approaches~\cite{10.1007/978-3-319-10602-1_41,Chang_2019_CVPR,Ding2018WeaklySupervisedAS,Huang2016ConnectionistTM,Li_2019_ICCV,Li_2020_CVPR,8578725,Richard_2018_CVPR,CrossTask}, we neither use the number of key-steps required to perform the task nor an ordered or unordered list, as it requires viewing the videos or defining heuristics, leading to scalability issues~\cite{multi-task-procl,joint_dynamic_summary}.
Additionally, learning various procedures requires numerous videos and annotating all the videos would consume considerable resources.
Motivated by this, we create \methsp as a self-supervised framework for procedure learning to create a scalable and efficient solution.

\mypara{Multimodal Procedure Learning}
Another class of methods work with multi-modal data, like narrated text and videos~\cite{Inria_dataset,BMVC.28.30,Doughty_2020_CVPR,Fried2020LearningTS,Malmaud2015WhatsCI,Sener_2015_ICCV,Shen_action_segmentation_2021_CVPR,10.1145/2647868.2654997,zhukov20}.
These works use Automatic Speech Recognition (ASR) to obtain the text, which is not perfect.
Due to this, the output needs to be manually cleaned, which is not scalable.
Additionally, such methods assume an alignment between the text and videos~\cite{Inria_dataset,Malmaud2015WhatsCI,10.1145/2647868.2654997}, which might not be accurate for most cases~\cite{multi-task-procl,joint_dynamic_summary}. 
Instead, we use only the visual modality as an input to the framework.
Due to this, we eliminate the need to obtain narrations that might be inaccurate and make our framework scalable.

\mypara{Learning Key-step Ordering}
Current works do not capture different key-step ordering to perform the same task.
They either assume a strict ordering~\cite{joint_dynamic_summary,kukleva2019unsupervised,VidalMata_2021_WACV} or do not predict the order~\cite{multi-task-procl,Shen_action_segmentation_2021_CVPR}.
However, we observe that subjects perform the same task in multiple ways (\Cref{fig:correspondences}), motivating us to capture different ways to accomplish the task.
Therefore, the final stage of \methsp aims to create a key-step order for each video and infer the relevant ordering to perform the task.

\mypara{Representation Learning for Procedure Learning}
Existing works on procedure learning employ various ways to create frame-wise features. To learn the representation space, Kukleva \etal~\cite{kukleva2019unsupervised} use relative timestamps of frames, and Vidal \etal~\cite{VidalMata_2021_WACV} predict the representation and timestamps of the future frames.
On the other hand, Elhamifar \etal either use the latent states obtained from an HMM~\cite{joint_dynamic_summary} or discover and utilize attention features from individual frames~\cite{multi-task-procl}. 
However, these methods do not exploit the signal provided by temporal correspondences, which is crucial for procedure learning, as we show in this work.

\mypara{Self-Supervised Representation Learning}
Learning a representation space without annotations saves substantial time and energy when creating deep learning solutions.
Motivated by this, recent works explore various pretext tasks to generate supervision signals for training deep learning architectures~\cite{Carreira2017QuoVA,resnet,Tran2015LearningSF,Tran2018ACL,Wang2018NonlocalNN}.
A few pretext tasks for learning image representations include image colourization~\cite{larsson2016learning,Larsson2017ColorizationAA}, object counting~\cite{Liu2018LeveragingUD,Noroozi2017RepresentationLB}, solving jigsaw puzzles~\cite{Carlucci2019DomainGB,Kim2018LearningIR}, predicting image rotations~\cite{8953870,komodakis2018unsupervised}, and reconstructing input images~\cite{10.5555/2987189.2987190} from noise~\cite{10.1145/1390156.1390294}.
Pretext tasks for learning video representations include predicting future frames~\cite{Ahsan2018DiscrimNetSA,Diba2019DynamoNetDA,Han19dpc,Kim2019SelfSupervisedVR,10.5555/3045118.3045209,10.5555/3157096.3157165}, using temporal order and coherence as labels~\cite{Fernando2017SelfSupervisedVR,Lee2017UnsupervisedRL,Misra2016ShuffleAL,Choi2020ShuffleAA,Xu_2019_CVPR}, and predicting the arrow of time~\cite{8578938}.

Video representation learning methods mentioned above employ a single video.
However, we want to identify similar key-steps in multiple videos for procedure learning.
To this end, we build upon existing video alignment techniques~\cite{tcc,lav} and devise a loss function that works well for procedure learning.
Note that procedure learning aims to find key-steps across a given set of videos; hence, it differs from video alignment.

\begin{table}[!thpb]
    \centering
    \caption{Comparison of datasets for Procedure Learning.
    The average number of key-steps and video length for \dataset are the highest, highlighting the complexity of the procedures included in \dataset}
    \setlength{\tabcolsep}{5pt}
    \begin{adjustbox}{width=1\textwidth}
    \begin{tabular}{@{}lccccc@{}}\toprule
    Dataset & Egocentric View & Manually Created & Avg. key-steps & Avg. Video Length (sec) & \#tasks\\
    \midrule
    Breakfast~\cite{Breakfast} & \textcolor{red}{\xmark} & \textcolor{green}{\cmark} & $5.1$ & $137.5$ & $10$ \\
    Inria~\cite{Inria_dataset} & \textcolor{red}{\xmark} & \textcolor{red}{\xmark} & $7.1$ & $178.8$ & $5$ \\
    ProceL~\cite{multi-task-procl} & \textcolor{red}{\xmark} & \textcolor{red}{\xmark} & $8.3$ & $251.5$ & $12$ \\
    CrossTask~\cite{CrossTask} & \textcolor{red}{\xmark} & \textcolor{red}{\xmark} & $7.4$ & $297$ & $\mathbf{18}$ \\
    \dataset (ours) & \textcolor{green}{\cmark} & \textcolor{green}{\cmark} & $\mathbf{8.7}$ & $\mathbf{769.2}$ & $16$\\
    \bottomrule
    \end{tabular}
    \end{adjustbox}
    \label{tab:dataset_stats}
\end{table}

\section{\dataset: Egocentric Video Dataset for Procedure Learning}
The \dataset dataset focuses on the key-steps required to perform a task instead of every action in the video.
To construct \dataset, we take two approaches: 
\begin{enumerate*}[label=(\alph*)]
\item identifying publicly available datasets that we annotate for key-steps;
\item recording new tasks to expand the range of tasks.
\end{enumerate*}
We follow the following criteria to shortlist from the public datasets:
\begin{enumerate*}[label=(\arabic*)]
\item The task should require multiple key-steps to perform. For example, preparing a sandwich involves a minimum of four key-steps~\cite{CMU_Kitchens}.
\item Videos of the same task must contain a similar set of key-steps. However, the order of the key-steps can differ.
\item To compare the performance of \methsp in egocentric and third-person views, we require a dataset with recordings of the same task in both views.
\item We prefer longer videos with sparse key-steps to generate practical solutions.
\end{enumerate*}

We select CMU-MMAC~\cite{CMU_Kitchens}, EGTEA Gaze+~\cite{egtea_gaze_p}, MECCANO~\cite{meccano}, and EPIC-Tents~\cite{tent} based on the above criteria.
CMU-MMAC contains recordings of subjects performing the same task from one egocentric and four third-person views.
Therefore, by using it, we compare the performance of \meth between egocentric and third-person views.
Though these four datasets include a diverse range of tasks, they do not contain tasks where the subject works in a constrained environment and deals with small objects (\eg, screws).
To alleviate this, we include manually recorded videos of assembling and disassembling a Personal Computer (PC).
This addition makes the dataset diverse and challenging in terms of variability in the size of objects involved and the complexity of key-steps (\eg, fixing the motherboard requires fastening nine screws).

\datasetsp contains videos and key-step annotations for multiple tasks from CMU-MMAC~\cite{CMU_Kitchens} and EGTEA Gaze+~\cite{egtea_gaze_p} and individual tasks like toy-bike assembly~\cite{meccano}, tent assembly~\cite{tent}, PC assembly, and PC disassembly.
\dataset consists of $62$ hours of annotated egocentric videos, including $16$ tasks with an average duration of $13$ minutes.
To annotate the videos for key-steps, we create a list of key-steps for each task, \eg, assembling a PC requires `Fix motherboard', `Fix hard disk', ..., `Place the cabinet cover'.
We use ELAN~\cite{elan} to annotate each video by marking the start and end location during which the key-step occurs.

\begin{figure}[t]
    \centering
    \includegraphics[width=0.9\textwidth]{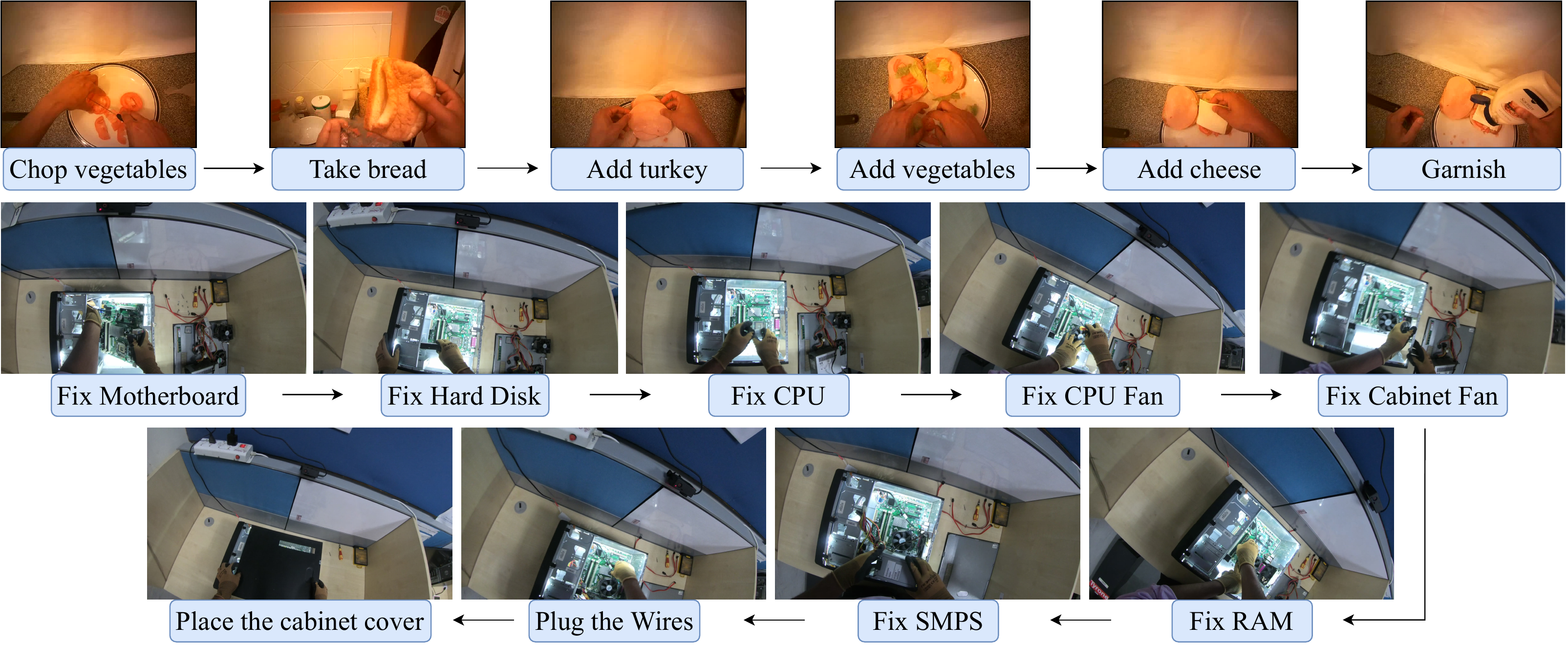}
    \caption{\textbf{Key-step annotations} for making turkey sandwich~\cite{egtea_gaze_p} and assembling a PC.
    }
    \label{fig:annotations}
\end{figure}

Along with various procedure learning tasks, \dataset is appropriate for understanding hand-object interaction, action forecasting and recognition, and a shared study of videos and text.
\Cref{fig:annotations} shows some example annotations and \Cref{tab:dataset_stats} compares \dataset with existing datasets.

We also considered a few other datasets that did not satisfy the requirements mentioned above~\cite{Sigurdsson2018CharadesEgoAL,Damen2018EPICKITCHENS}. The reasons for their non-inclusion are given in the supplementary.

\section{Correspond and Cut Framework for Procedure Learning}
\begin{figure*}[t]
    \centering
    \includegraphics[width=\textwidth]{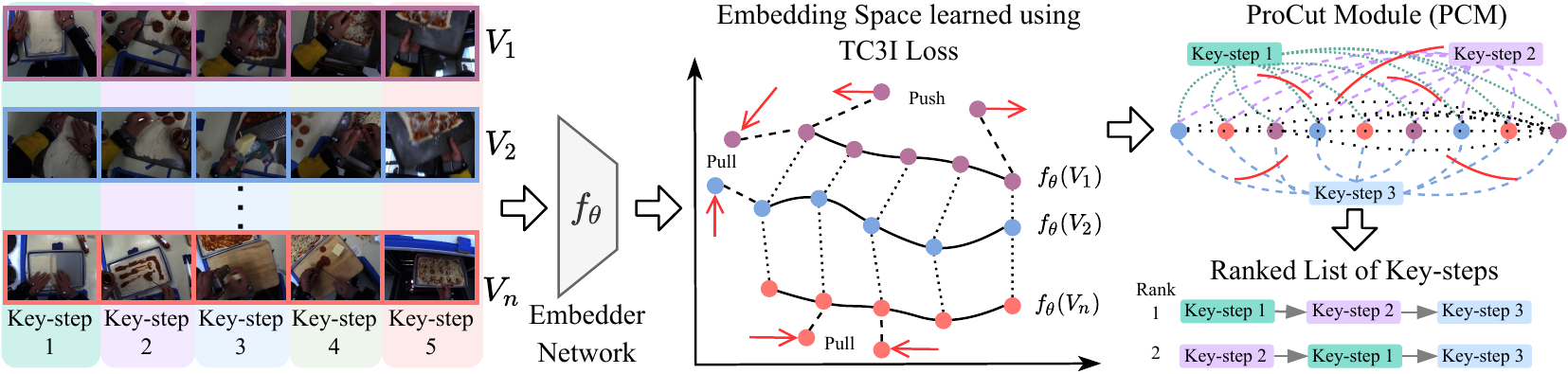}
    \caption{\textbf{Correspond and Cut (\meth) framework for Procedure Learning.}
    \methsp takes in multiple videos from the same task and passes them through the embedder network trained using the proposed TC3I loss.
    The goal of the embedder network is to learn similar embeddings for corresponding key-steps from multiple videos and for temporally close frames.
    The ProCut Module (PCM) localizes the key-steps required for performing the task.
    PCM converts the clustering problem to a multi-label graph cut problem solved using the Alpha Expansion algorithm~\cite{alpha_expansion}.
    The output provides the assignment of frames to the respective key-steps and their ordering.
    }
    \label{fig:methodology}
\end{figure*}

Humans often follow the same steps to perform any particular task, though the order of steps might be different.
This work proposes a methodology which, given a set of videos of humans performing a task, learns similar embeddings across videos for the key-steps required to complete a task.
Once we have the embeddings, learning a procedure reduces to clustering the embeddings for localizing the key-steps among all the videos.
To learn the embeddings, we exploit temporal correspondences between the videos of the same task.
For that purpose, we train a representation learning network using the proposed TC3I loss.
TC3I builds on top of existing temporal video alignment methods~\cite{tcc,lav}.
After learning the embeddings, we use PCM, shown in \Cref{fig:methodology}, to cluster and localize the underlying key-steps.
PCM models the clustering problem as a multi-label graph cut problem and solves it to localize the key-steps.
Once we localize the key-steps using PCM, we use the frame's relative location in a video to generate the key-step ordering for each video.

\mypara{Notation} 
\methsp takes in $V=\{V_i: i \in \mathbb{N}, 1 \leq i \leq n\}$ untrimmed videos of the same task, where $n$ is the total number of videos.
Each of the $n$ videos can have a different number of frames.
We denote the embedding function used to generate the frame-level embeddings as $f_{\theta}$, which is a neural network with parameters $\theta$.
A video $V_k$ with $m$ frames is denoted as $V_k = \{f_k^1, f_k^2, \dots, f_k^m\}$ and the video's frame-level embeddings are denoted as $f_{\theta}(V_k) = \{v_k^1, v_k^2, \dots, v_k^m\}$.
We assume $K$ key-steps in a task, where $K$ is a hyper-parameter.

\subsection{Learning the Embeddings using the TC3I loss}
\label{sec:learning_representation_space}
We aim to learn similar embeddings for the frames with comparable semantic information across different temporal locations from multiple videos.
For that purpose, we use Temporal Cycle Consistency (TCC)~\cite{tcc} to find correspondences across time in videos.

Consider two videos $V_1$ and $V_2$, with lengths $p$ and $q$, respectively.
To check if a point $v_1^i$ in $V_1$ is cycle consistent, its nearest neighbour $v_2^j = \arg \min_{v_2 \in V_2} \lVert v_1^i - v_2 \rVert$ is calculated in $V_2$. Then the process is repeated for $v_2^j$ in $V_1$ to get $v_1^k = \arg \min_{v_1 \in V_1} \lVert v_2^j - v_1 \rVert$.
If $i=k$, then the point is considered cycle consistent.
An acceptable embedding space consists of a maximum number of cycle-consistent points for a pair of sequences.
Specifically, for a point $v_1^i$ in $V_1$, we determine its soft nearest neighbor $\tv_2$ in $V_2$ by using the softmax function as follows:
\begin{equation}
    \tv_2 = \sum_j \alpha_j v_2^j, \quad \text{where} \quad 
    \alpha_j = \frac{e^{-\lVert v_1^i - v_2^j \rVert^2}}{\sum_k e^{- \lVert v_1^i - v_2^k \rVert^2}}.
\end{equation}
Here $\alpha_j$ signifies the \textit{similarity} between $v_1^i$ and individual $v_2^j \in V_2$.
Once we have the soft nearest neighbor, a similarity vector $\beta_1^i$ is calculated.
$\beta$ defines the proximity between $\tv_2$ and each frame $v_1^k \in V_1$ as:
\begin{equation}
    \beta_1^i[k] = \frac{e^{-\lVert \tv_2 - v_1^k \rVert^2}}{\sum_j e^{-\lVert \tv_2 - v_1^j \rVert^2}}.
\end{equation}
As $\beta$ is a discrete distribution of similarities over time, it peaks around the $i^{th}$ time index.
To avoid this, a Gaussian prior is applied to $\beta$ by minimizing the normalized squared distance $\frac{|i - \mu|^2}{\sigma^2}$ as the objective.
By applying additional variance regularization, $\beta$ is enforced to be peaky around $i$.
Hence, the final cycle consistency loss between videos $V_1$ and $V_2$, corresponding to $i^\text{th}$ frame of $V_1$ is:
\begin{equation}
    L(V_1, V_2, v_1^i) = \frac{|i - \mu|^2}{\sigma^2} + \lambda \log(\sigma).
\end{equation}
Here, $\mu_i =\sum_k \beta_1^i[k] \times k$ and $\sigma^2_i = \sum_k \beta_1^i[k] \times (k - \mu_i)^2$, and $\lambda$ is the regularization weight.
Formulating TCC in this way ensures the model is not heavily penalized when it cycles back to close-by frames.

We observe that there are many repetitive frames in egocentric videos because of which cycle consistency loss often loops back to similar but temporally far-away frames. 
To alleviate the issue, we utilize the Contrastive-Inverse Difference Moment (C-IDM) loss~\cite{lav} (a modified form of Inverse Difference Moment~\cite{idm}) for applying temporal regularization on each video.
The C-IDM loss between the two frames $i$ and $j$ of a video $V_1$ is computed as:
\begin{align}
    I(V_1,i,j) = \left( 1 - \mathcal{N}(i, j) \right) \gamma(i, j) \max \left( 0, \zeta - d(i, j) \right) + \mathcal{N}(i, j) \frac{d(i, j)}{\gamma(i, j)}.
\end{align}
Here, \mbox{$\gamma(i,j) = (i - j)^2 + 1$}, $d(i, j)$ is the Euclidean distance between $f_{\theta}(v_1^i)$, and $f_{\theta}(v_1^j)$, $\zeta$ is the margin parameter, and $\mathcal{N}$ is the neighborhood function such that, \mbox{$\mathcal{N}(i, j) = 1$} if $|i - j| \leq \sigma$, and $0$ otherwise. Here, $\sigma$ is the window size for separating temporally far away frames.
The C-IDM loss encourages the embeddings of the temporally close frames to be similar and the embeddings of temporally far frames to be dissimilar.
The final loss combines both TCC and C-IDM (referred to as TC3I loss from now on):
\begin{align}
    \text{TC3I}(V_1, V_2) & = \sum_{i \in V_1} L(V_1, V_2, v_1^i) 
    				 + \sum_{j \in V_2} L(V_1, V_2, v_2^j) \nonumber \\
    			   & + \xi \sum_{i \in V_1} \sum_{j \in V_1} I(V_1,i,j) 
    				 + \xi \sum_{i \in V_2} \sum_{j \in V_2} I(V_2,i,j).
\end{align}
Here, $\xi$ is a regularization parameter.

\subsection{Localizing the Key-Steps using the ProCut Module} 
\label{sec:keystep_identification}
\begin{figure}[t]
    \centering
    \includegraphics[width=\textwidth]{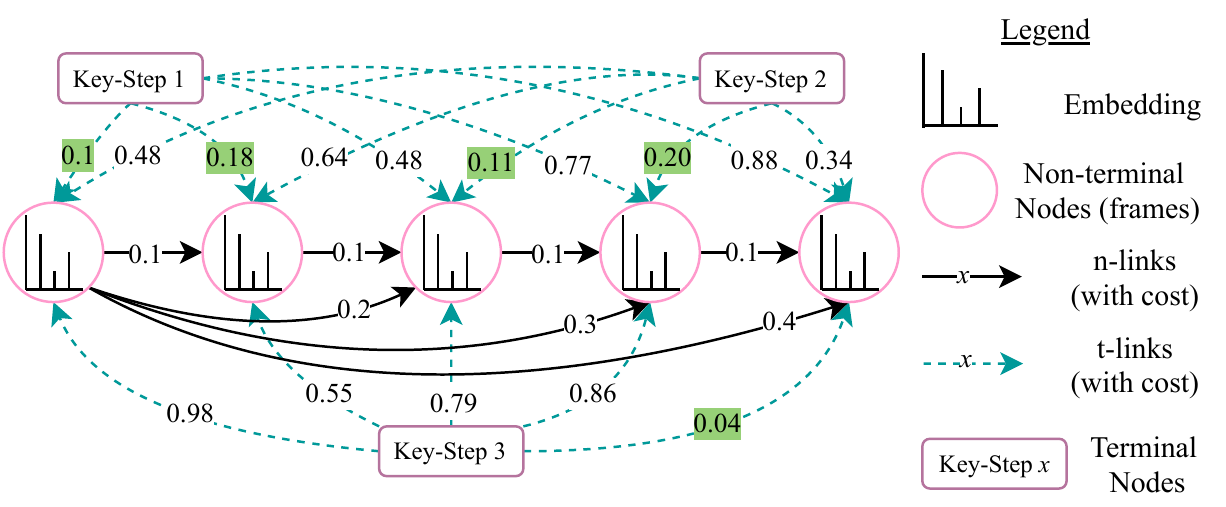}
    \caption{\textbf{ProCut Module (PCM).}
    Non-terminal nodes in the graph represent the embeddings of the frames.
    Terminal nodes represent the key-steps required to perform the task.
    The terminal and non-terminal nodes are connected using the t-links.
    Non-terminal nodes are connected using the n-links.
    The numbers inscribed in arrows represent the cost of using the respective link.
    Costs highlighted in green represent the lowest cost to assign a frame to the key-step.
    For brevity, n-links are shown only for the first non-terminal node.
    Diagram best viewed in colour.
	}
    \label{fig:graph_cut}
\end{figure}

Once we learn the embeddings, we aim to localize the key-steps required for performing the task.
Kukleva \etal~\cite{kukleva2019unsupervised} localize the key-steps by generating $K$ clusters of embeddings using the K-Means algorithm~\cite{KMeans}. However, they need to assume a fixed order of key-steps to assign frames to the key-steps.
Instead, we propose a novel ProCut Module (PCM) for the purpose. PCM converts the clustering problem to a multi-label graph cut problem~\cite{graph_cut}, as described below.

Let $G = \langle V, E \rangle$ be a graph consisting of a set of nodes $V$ and a set of directed edges $E$ connecting them.
The node set $V$ consists of $K$ \textit{terminal nodes} representing the key-steps, and \textit{non-terminal nodes} (equal to the number of frames) representing the embeddings of the frames generated using the Embedder network.
There are two kinds of edges in the graph: \textit{t-links} connecting non-terminal nodes to the terminal nodes, and \textit{n-links} connecting two non-terminal nodes.

We use the Fuzzy C-Means algorithm~\cite{fuzzy_c_means} to assign a cost to the \textit{t-links}.
The algorithm performs soft clustering and calculates the probability of a frame belonging to each cluster.
We subtract the probability value from $1$ to obtain the cost of assigning a frame to each cluster.
The cost value for the \textit{n-links} is assigned based on the temporal distance between the nodes.
For example, if the nodes are temporally closer (\eg, nodes at positions $1$ and $2$ in \Cref{fig:graph_cut}), the cost of assigning the same label to them is lower, otherwise (\eg, for nodes at positions $1$ and $5$ in \Cref{fig:graph_cut}), the cost is high.
After creating the graph $G$, we use $\alpha$-Expansion~\cite{alpha_expansion} to find the minimum cost cut.
We use the discovered cut to assign frames to $K$ labels. 
As shown in \Cref{fig:graph_cut}, the lowest costs (highlighted in green) result in assigning the first and second frames to key-step $1$, the third and fourth frames to key-step $2$, and the last to key-step $3$.

\subsection{Determining Order of the Key-Steps}
\label{sec:keystep_ordering}
When it comes to determining the ordering of the key-steps, it makes sense to allow each video to have a distinct key-step ordering as there can be multiple ways to perform a task.
However, current works either use a fixed order of key-steps to decode all the videos~\cite{joint_dynamic_summary,kukleva2019unsupervised,VidalMata_2021_WACV} or do not predict the ordering~\cite{multi-task-procl,Shen_action_segmentation_2021_CVPR}.
One of the advantages of using \methsp to determine the key-step is that it allows each video to have its independent order of the key-steps.

To infer the sequential order of key-steps, we calculate the normalized time for each frame $v_i^n$ in video $V_i$ consisting of $p$ frames as $T(v_i^n) = \frac{n}{p}$~\cite{kukleva2019unsupervised,VidalMata_2021_WACV}. Then we calculate the time instant for each cluster as the average normalized for frames assigned to it. 
The clusters are then arranged in increasing order of the average time, providing us with the sequence of key-steps used to perform the task in a video.
Once we have key-step order for all the videos of the same task, we generate their ranked list based on the number of times the subjects followed a particular order.
The order followed the most ends up being at the top of the ranked list.
Doing this enables us to determine different sequential orders of key-steps to accomplish a task.

\subsection{Implementation Details}
\label{sec:implementation_detail}
We use ResNet-50~\cite{resnet} as our backbone network to extract the features.
Motivated by~\cite{tcc}, for training the Embedder network, we use a pair of training videos at a time, select frames at random within the videos and optimize the proposed TC3I loss until convergence.
The features are extracted from the \textit{Conv4c} layer and a stack of $c$ context frames features is created along the temporal dimension.
We reshape our input video frames to $224 \times 224$.
To aggregate the temporal information, we pass the combined features through two 3D convolutional layers followed by a 3D global max pooling layer, two fully-connected layers, and a linear projection layer to output the embeddings of dimension $128$.
We set the value of $K$ to $7$ and compare the performance of \methsp with the other values of $K$ in \Cref{tab:selecting_k}.
Furthermore, for all our experiments, we follow the task-specific settings laid out in~\cite{multi-task-procl}.
We use PyTorch~\cite{PyTorch} for all our experiments.

\section{Experiments} 
\label{sec:experiments}

\subsection{Evaluation} \label{subsec:evaluation}
Current works compute framewise F1-Score and IoU scores for key-step localization~\cite{multi-task-procl,joint_dynamic_summary,kukleva2019unsupervised,Shen_action_segmentation_2021_CVPR,VidalMata_2021_WACV}.
The F1-Score is a harmonic mean of precision and recall scores.
For calculating recall, the ratio between the number of frames having correct key-steps prediction and the number of ground truth key-step frames across all the key-steps of a video is calculated.
For precision, the denominator is the number of frames assigned to the key-steps.
For calculations, the one-to-one mapping between the ground truth and prediction is generated using the Hungarian algorithm~\cite{hungarian_algorithm} following~\cite{Inria_dataset,multi-task-procl,joint_dynamic_summary,Shen_action_segmentation_2021_CVPR,kukleva2019unsupervised}.
However, these metrics tend to assign high scores to models that assign most frames to a single cluster, as the key-step with most frames matches with the background frame's label in the ground truth.
Furthermore, for untrimmed procedure learning videos, most of the frames are background, resulting in high scores.

Shen \etal~\cite{Shen_action_segmentation_2021_CVPR} attempt to solve this problem by analyzing the MoF score, but as pointed out in~\cite{kukleva2019unsupervised}, MoF is not always suitable for an imbalanced dataset.
Instead, we propose calculating the framewise scores for each key-step separately and then taking the mean of the scores over all the key-steps.
This penalises the cases when there is a large performance difference for different key-step, \eg, when all the frames get assigned to a single key-step.
Upon following this protocol, the scores for all the methods decrease.
This paper presents the results generated using the proposed evaluation protocol unless otherwise mentioned.

\subsection{Procedure Learning from Third-person Videos}
\begin{table*}[t]
    \centering
    \caption{\textbf{Procedure Learning from Third-person Videos}. Comparison between state-of-the-art methods and \methsp on benchmark third-person video datasets~\cite{joint_dynamic_summary,CrossTask}. Our method outperforms all the techniques using videos only (in F-Score). It even manages to give at par performance compared to the techniques using multi-modal input. \textbf{P}, \textbf{R}, and \textbf{F} represent precision, recall, and F-score respectively}
    \setlength{\tabcolsep}{5pt}
    \begin{tabular}{@{}lcrrrrrr@{}}\toprule
        & \multirow{2}{*}{Input Modality} & \multicolumn{3}{c}{ProceL~\cite{joint_dynamic_summary}} & \multicolumn{3}{c}{CrossTask~\cite{CrossTask}}\\
        \cmidrule(lr){3-5}\cmidrule(l){6-8}
        & & \textbf{P} & \textbf{R} & \textbf{F} & \textbf{P} & \textbf{R} & \textbf{F} \\
        \midrule
        Uniform & Video & $12.4$ & $9.4$ & $10.3$ & $8.7$ & $9.8$ & $9.0$\\
        Alayrc \etal~\cite{Inria_dataset} & Video + Narrations & $12.3$ & $3.7$ & $5.5$ & $6.8$ & $3.4$ & $4.5$ \\
        Kukleva \etal~\cite{kukleva2019unsupervised} & Video & $11.7$ & $30.2$ & $16.4$ & $9.8$ &  $35.9$& $15.3$ \\
        Elhamifar \etal~\cite{multi-task-procl} & Video & $9.5$ & $26.7$ & $14.0$ & $10.1$ & $\mathbf{41.6}$ & $16.3$ \\
        Fried \etal~\cite{Fried2020LearningTS} & Video & $-$ & $-$ & $-$ & $-$ & $28.8$ & $-$ \\
        Shen \etal~\cite{Shen_action_segmentation_2021_CVPR} & Video + Narrations & $16.5$ & $\mathbf{31.8}$ & $21.1$ & $15.2$ & $35.5$ & $21.0$  \\
        \methsp (\textit{ours}) & Video & $\mathbf{20.7}$ & $22.6$ & $\mathbf{21.6}$ & $\mathbf{22.8}$ & $22.5$ & $\mathbf{22.6}$ \\
        \bottomrule
    \end{tabular}
    \label{tab:table_third_person_results}
\end{table*}

To test the generalizability of \meth on third-person videos and to ensure a fair comparison with existing methods~\cite{Inria_dataset,multi-task-procl,Fried2020LearningTS,kukleva2019unsupervised,Shen_action_segmentation_2021_CVPR}, we perform experiments on third-person procedure learning benchmark datasets: ProceL~\cite{joint_dynamic_summary} and CrossTask~\cite{CrossTask}. We obtain the results of previous works from~\cite{Shen_action_segmentation_2021_CVPR}. Note that here we use the evaluation protocol employed by the previous works~\cite{multi-task-procl,joint_dynamic_summary,kukleva2019unsupervised,Shen_action_segmentation_2021_CVPR}.
As seen in \Cref{tab:table_third_person_results}, \methsp outperforms other methods (in terms of the F-Score) utilizing only videos as the input modality.
Further, with only video as the input modality, \methsp even manages to perform at par with multi-modal methods.
Previous works have used different forms of self-supervision.
For example,~\cite{multi-task-procl} use the pseudo-labels provided by subset selection and~\cite{kukleva2019unsupervised} utilize the relative time-stamps of video frames.
Instead, the comparison in \Cref{tab:table_third_person_results} shows that the signal provided by corresponding frames is superior for the task of procedure learning.

\subsection{Procedure Learning Results from Egocentric Videos}
\mypara{Baselines}
We consider three baseline methods:
\begin{enumerate}
    \item \textbf{Random.} Here we predict the labels by randomly sampling predictions from a uniform distribution with $K$ values representing $K$ key-steps.
    \item \textbf{TC3I + HC.} Instead of PCM, we use the K-Means algorithm and generate $K$ clusters from the representation space.
    \item \textbf{TC3I + SS.} Here, instead of PCM, we use subset selection for the key-step assignment. The algorithm takes in the frame's embeddings and $M$ (hyper-parameter) latent states obtained using K-Means~\cite{KMeans}.
It then selects a subset $S$ (of size $K$) of the states as key-steps and finds the frames' assignments.
We use the greedy algorithm used in~\cite{multi-task-procl} to perform subset selection. Refer to the supplementary material for the hyper-parameter values.
\end{enumerate}

\begin{figure}[t]
    \centering
    \includegraphics[width=\textwidth]{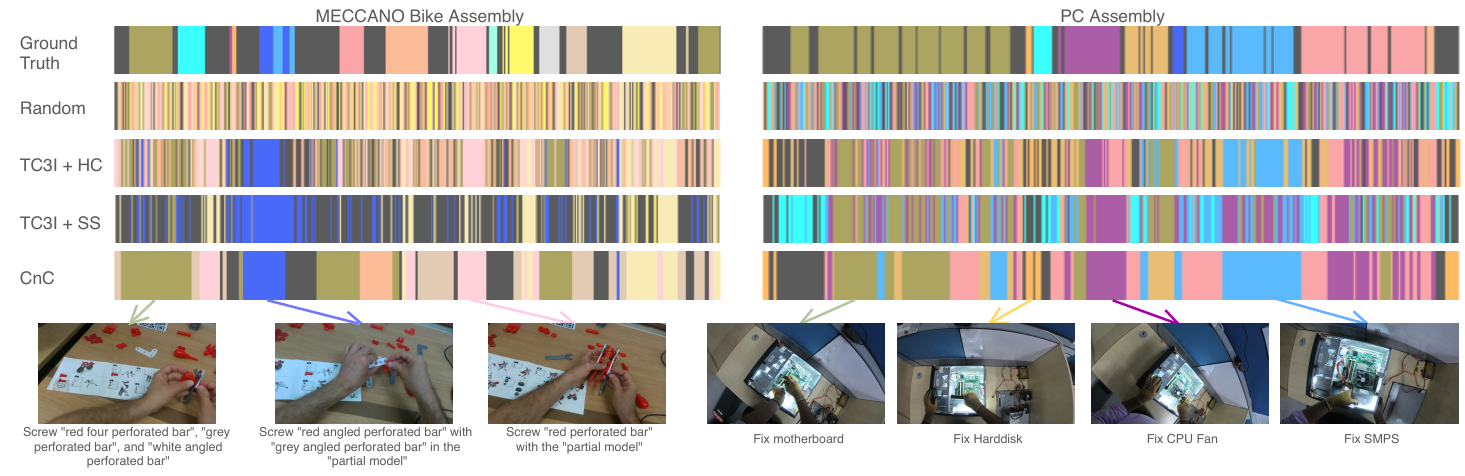}
    \caption{\textbf{Qualitative results} for MECCANO and PC Assembly highlight the effectiveness of \meth. Additionally, PCM outperforms HC and SS when clustering the key-steps. Furthermore, due to the TC3I loss, CnC correctly identifies the key-steps that are short (fix a hard disk in PC Assembly). The gray segments denote the background.
    }
    \label{fig:qualitative_results}
\end{figure}

\Cref{tab:results_procedure_learning} summarises the results obtained on \datasetsp using the baselines and proposed \meth.
\meth performs higher than all the three baselines. This is due to (a) the ability of the TC3I loss to learn the representation space where similar key-steps lie close without enforcing any ordering or temporal constraints.
Moreover, TC3I adds temporal coherency to the learned representations by adopting the C-IDM loss~\cite{lav} (\Cref{fig:qualitative_results}).
(b) PCM gains a comprehensive view of the problem by considering the cost of assigning each frame belonging to every key-step and its temporal relationship with the other frames.
\begin{table*}[t]
    \centering
    \caption{\textbf{Procedure Learning Results} obtained on \dataset. Here, \methsp performs the best, highlighting the effectiveness of the TC3I loss and PCM
    }
    \begin{adjustbox}{width=1\textwidth}
    \begin{tabular}{@{}lcccccccccccc@{}}
    \toprule
    & \multicolumn{12}{c}{\dataset}\\
    \cmidrule{2-13}
    & \multicolumn{2}{c}{CMU-MMAC} & \multicolumn{2}{c}{EGTEA G.} & \multicolumn{2}{c}{MECCANO} & \multicolumn{2}{c}{EPIC-Tents} & \multicolumn{2}{c}{PC Assembly} & \multicolumn{2}{c}{PC Disas.} \\
    \cmidrule(lr){2-3}\cmidrule(lr){4-5}\cmidrule(lr){6-7}\cmidrule(lr){8-9}\cmidrule(lr){10-11}\cmidrule(l){12-13}
    & F1 & IoU & F1 & IoU & F1 & IoU & F1 & IoU & F1 & IoU & F1 & IoU \\
    \midrule
    Random & $15.7$ & $5.9$ & $15.3$ & $4.6$ & $13.4$ & $5.3$ & $14.1$ & $6.5$ & $15.1$ & $7.2$ & $15.3$ & $7.1$\\
    TC3I + HC & $19.2$ & $9.0$ & $20.8$ & $7.9$ & $16.6$ & $\mathbf{8.0}$ & $15.4$ & $7.8$ & $21.7$ & $11.0$ & $24.9$ & $14.1$\\
    TC3I + SS & $19.7$ & $8.9$ & $20.4$ & $7.9$ & $16.3$ & $7.8$ & $15.9$ & $7.8$ & $24.8$ & $11.9$ & $23.6$ & $14.0$\\
    \methsp & $\mathbf{22.7}$ & $\mathbf{11.1}$ & $\mathbf{21.7}$ & $\mathbf{9.5}$ & $\mathbf{18.1}$ & $7.8$ & $\mathbf{17.2}$ & $\mathbf{8.3}$ & $\mathbf{25.1}$ & $\mathbf{12.8}$ & $\mathbf{27.0}$ & $\mathbf{14.8}$\\
    \bottomrule
    \end{tabular}
    \end{adjustbox}
    \label{tab:results_procedure_learning}
\end{table*}

\methsp performs better on long sequences as the TC3I loss compensates by searching for corresponding frames in the entire length of the videos, making it possible to learn a reasonable representation space despite the length of the videos.

Further, the results in \Cref{tab:results_procedure_learning} show that PCM is superior for key-frame clustering and assignment along with TC3I as it results in the highest F-Score and IoU on \dataset.
The gain in performance is because PCM considers the cost of assigning each frame to every key-step and its temporal relationship with the other frames (\Cref{fig:qualitative_results}).
This allows PCM to gain a comprehensive view of the problem compared to HC, which does not consider the cost of each frame belonging to other key-steps and SS, which has lower generalisation power~\cite{multi-task-procl}.

\subsection{Egocentric vs. Third-person Videos}
\setlength{\intextsep}{0cm}
\begin{wraptable}{O}{5cm}
    \centering
    \caption{\textbf{Egocentric vs. Third-person results.} We use different views from~\cite{CMU_Kitchens} for comparison.
    We obtain better results using \meth on egocentric videos highlighting their effectiveness.
    \textbf{P}, \textbf{R}, and \textbf{F} denote precision, recall, and F-score respectively
    }
    \begin{tabular}{@{}lrrrr@{}}\toprule
        \textbf{View} & \textbf{P} & \textbf{R} & \textbf{F} & \textbf{IoU} \\
        \midrule
        TP (Top)    & $17.4$ & $18.4$ & $17.9$ & $8.1$ \\
        TP (Back)   & $18.8$ & $21.5$ & $20.0$ & $8.5$ \\
        TP (LHS)    & $21.2$ & $22.7$ & $21.8$ & $9.7$ \\
        TP (RHS)    & $19.8$ & $21.7$ & $20.6$ & $8.7$ \\
        Egocentric  & $\mathbf{21.6}$ & $\mathbf{24.4}$ & $\mathbf{22.7}$ & $\mathbf{11.1}$ \\
        \bottomrule
    \end{tabular}
    \label{tab:first_person_vs_third_person}
\end{wraptable}

Here, we compare the results obtained after training \methsp on multiple views from CMU-MMAC~\cite{CMU_Kitchens}.
As seen in \Cref{tab:first_person_vs_third_person}, the frame-wise F1-Score and IoU scores are the highest for the egocentric view.
This is because egocentric videos offer lower occlusion by the expert's body and provide higher visibility of hand-object interactions.
This highlights one of the central hypotheses of this paper: the effectiveness of using egocentric videos over third-person videos for procedure learning.
Also, we observe that the results vary for third-person videos due to the camera placement.
This increases one variable when creating data for procedure learning. 
Alternatively, egocentric videos use head-mounted cameras, eliminating uncertainty.

\subsection{Ablation study}
\label{subsec:ablation}
Here, we quantitatively evaluate our design choices.
Due to space constraints, results for~\cite{CMU_Kitchens} and~\cite{egtea_gaze_p} are provided here, and the rest are in the supplementary.
\begin{table}[!tphb]
    \centering
    \caption{\textbf{Effectiveness of the TC3I loss.}
    TC3I loss outperforms other losses as it focuses on corresponding frames and employs C-IDM for temporal coherency
    }
	\setlength{\tabcolsep}{7pt}
    \begin{tabular}{@{}lrrrrrr@{}}\toprule
        \multirow{2}{4em}{Experiment} & \multicolumn{3}{c}{CMU-MMAC~\cite{CMU_Kitchens}} & \multicolumn{3}{c}{EGTEA Gaze+~\cite{egtea_gaze_p}}\\
        \cmidrule(lr){2-4}\cmidrule(l){5-7}
        & Precision & F-Score & IoU & Precision & F-Score & IoU \\
        \midrule
        TCC + PCM   & $18.5$ & $19.7$ & $9.5$ & $17.5$ & $19.7$ & $8.8$ \\
        LAV + TCC + PCM & $18.8$ & $19.7$ & $9.0$ & $16.4$ & $18.6$ & $7.5$\\
        LAV + PCM & $20.6$ & $21.1$ & $9.4$ & $17.4$ & $19.1$ & $8.0$\\
        TC3I + PCM (\meth) & $\mathbf{21.6}$ & $\mathbf{22.7}$ & $\mathbf{11.1}$ & $\mathbf{19.6}$ & $\mathbf{21.7}$ & $\mathbf{9.5}$ \\\bottomrule
    \end{tabular}
    \label{tab:results_ablation_tc3i}
\end{table}

\subsubsection{Effectiveness of the TC3I Loss:}
Here, we replace the TC3I loss in \methsp with TCC~\cite{tcc}, LAV~\cite{lav}, and a combination of LAV and TCC~\cite{lav} to study the efficacy of the proposed TC3I loss.
TC3I loss in \Cref{tab:results_ablation_tc3i} obtains the highest F-Score and IoU.
As observed in our initial set of experiments, TCC loss lacks temporal coherency due to which temporally close frames do not lie close in the learned representation space, resulting in lower results when compared to TC3I and LAV, which account for temporal coherency using the C-IDM loss.
For LAV + TCC, our observations are consistent with~\cite{lav} because there is no performance gain when directly combining LAV and TCC losses since LAV works on L2-normalised embeddings, whereas TCC does not~\cite{lav}.
The LAV loss performs better than TCC and LAV + TCC; however, the results are not better than TC3I because the Soft-DTW used in LAV accounts for global alignment.
However, LAV does not focus on the per-frame features~\cite{lav}, which is beneficial when looking for similar key-steps in different videos.
The TC3I loss overcomes these issues by focusing on correspondences in multiple videos at frame level and adding temporal coherency by adopting the C-IDM loss.

\subsubsection{Selecting the value of $K$}
\Cref{tab:selecting_k} contains results of \meth and the baselines as the function of $K$.
Additionally, it features the results after replacing PCM with HC and SS as the function of $K$.
Here, key observations are: (a) \methsp performs the best when $K=7$,
(b) the results do not change significantly for \methsp as $K$ increases.
However, we observe a decline in the results for HC and SS as K increases, highlighting the effectiveness of PCM for key-step localisation.
\begin{table}[t]
    \centering
    \caption{\textbf{Selecting $K$.} Results with various values of $K$.
    Numbers in bold are highest in the respective row, and underlined numbers are highest in the respective column}
	\setlength{\tabcolsep}{6.6pt}
    \begin{tabular}{@{}lrrrrrrrr@{}}\toprule
        \multirow{2}{4em}{Experiment} & \multicolumn{4}{c}{CMU-MMAC~\cite{CMU_Kitchens}} & \multicolumn{4}{c}{EGTEA Gaze+~\cite{egtea_gaze_p}} \\
        \cmidrule(l){2-5}\cmidrule(l){6-9}
        &   $K$=$7$   &  $K$=$10$   &  $K$=$12$   &  $K$=$15$ &   $K$=$7$   &  $K$=$10$   &  $K$=$12$   &  $K$=$15$  \\
        \midrule
        Random & $\mathbf{15.7}$ & $12.7$ & $11.6$ & $10.4$ & $\mathbf{15.4}$ & $12.3$ & $11.4$ & $10.4$ \\
        TC3I + HC & $\mathbf{19.2}$ & $17.4$ & $16.3$ & $16.8$ & $\mathbf{20.8}$ & $17.8$ & $16.7$ & $17.3$\\
        TC3I + SS & $\mathbf{19.7}$ & $17.3$ & $17.0$ & $15.7$  & $\mathbf{20.4}$ & $17.8$ & $16.7$ & $16.8$ \\
        \methsp & $\mathbf{\underline{22.7}}$ & $\underline{19.1}$ & $\underline{20.4}$ & $\underline{20.1}$ & $\mathbf{\underline{21.7}}$ & $\underline{19.9}$ & $\underline{19.9}$ & $\underline{19.9}$ \\
        \bottomrule
    \end{tabular}
    \label{tab:selecting_k}
\end{table}

\section{Conclusion}
Learning procedures from the visual demonstration of a task by an expert, is an important step in scaling the learning capabilities of autonomous agents. Unlike current state-of-the-art techniques, instead of third-person videos, we have proposed procedure learning from first-person viewpoint. Given the unavailability of the datasets for the purpose, we proposed the \dataset containing egocentric videos for procedure learning. We also proposed a new technique, \meth, for procedure learning from egocentric videos that utilize the proposed TC3I loss to learn an embedding space in a self-supervised fashion. Finally, we employ PCM to identify the key-steps. Our results demonstrate the superiority of using the egocentric view and the effectiveness of the proposed technique for procedure learning.

\paragraph{Acknowledgements.}
The work was supported in part by the Department of Science and Technology, Government of India, under DST/ICPS/Data-Science project ID T-138. We acknowledge Pravin Nagar and Sagar Verma for sharing the PC Assembly and Disassembly videos recorded at IIIT Delhi. We also acknowledge Jehlum Vitasta Pandit and Astha Bansal for their help with annotating a portion of EgoProceL.

\section{Appendix}
Additional details to support the main paper.
\begin{figure}[!tphb]
    \centering
    \includegraphics[width=\textwidth]{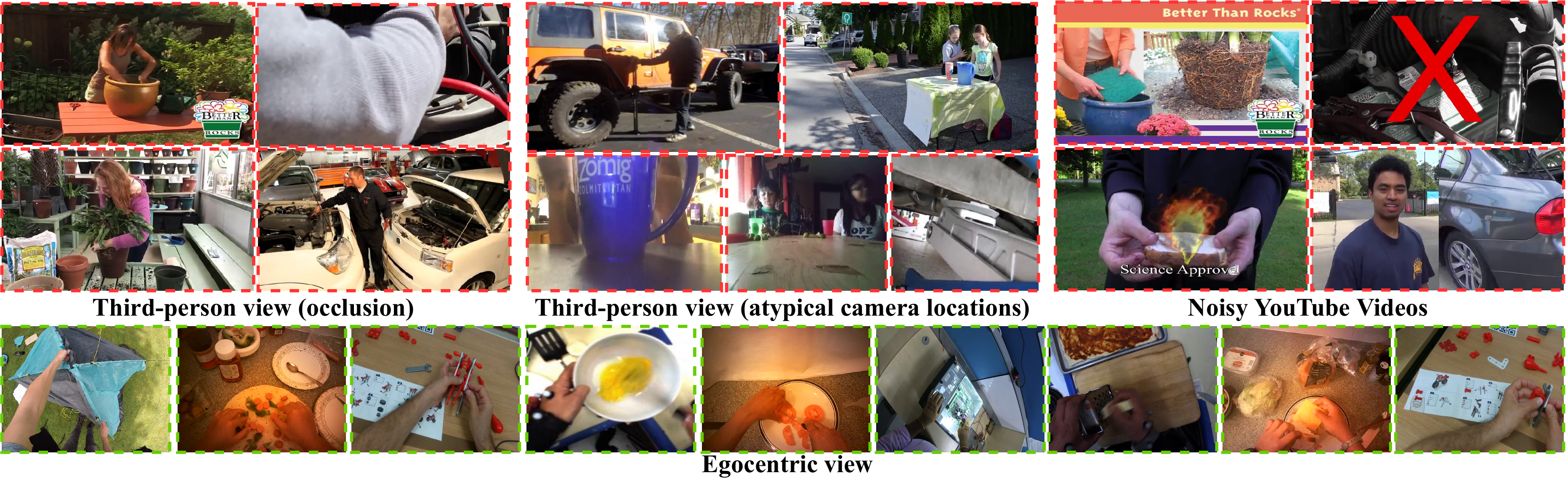}
    \caption{\textbf{Issues with standard datasets for procedure learning.}
    Existing datasets~\cite{Inria_dataset,joint_dynamic_summary,Breakfast,howto100m,COIN,YouCook2,CrossTask} majorly consist of third-person videos. 
    They contain issues like occlusion and atypical camera locations that make them ill-suited for procedure learning.
    Additionally, the datasets rely on noisy videos from YouTube~\cite{joint_dynamic_summary,howto100m,COIN,CrossTask}.
    In contrast, we propose to use egocentric videos that overcome the issues posed by third-person videos.
    To this end, we create the \dataset dataset.
    }
    \label{fig:first_person_vs_third_person}
\end{figure}

\subsection{Outline}
\Cref{fig:first_person_vs_third_person} highlights issues with standard third-person datasets, motivating us to use egocentric videos for procedure learning.
In \Cref{sec:dataset_details}, we discuss the annotation protocols, task-level details, and datasets excluded while creating the \dataset dataset.
In \Cref{sec:applications}, we highlight multiple use-cases for our work.
In \Cref{subsec:egtea_ablation_results}, we provide additional ablation results on \dataset.
To facilitate reproducing the results reported in the main paper and supplementary, \Cref{subsec:hyper-parameters} lists the hyper-parameters used for \meth.
Furthermore, we release the \dataset dataset and code for the work on project's webpage\footnote{\label{project_page}Link 1: \href{http://cvit.iiit.ac.in/research/projects/cvit-projects/egoprocel}{http://cvit.iiit.ac.in/research/projects/cvit-projects/egoprocel}; Mirror link 2: \href{https://sid2697.github.io/}{https://sid2697.github.io/}}.

\section{\dataset}\label{sec:dataset_details}
This section contains additional details on the proposed \dataset dataset.

\subsection{Annotation Protocols followed for \dataset}\label{subsec:annotation}
\paragraph{CMU-MMAC~\cite{CMU_Kitchens}, EPIC-Tents~\cite{tent}, MECCANO~\cite{meccano}, PC Assembly, PC Disassembly:} A list of key-steps required to perform the task was created upon viewing the videos.
Two annotators were asked to identify the key-steps in the videos and temporally mark the start and end locations.
Once an annotator added temporal segments to the videos, the other annotator verified them.
We use the ELAN software~\cite{elan} to annotate the videos.

\paragraph{EGTEA Gaze+~\cite{egtea_gaze_p}:} We used the recipes provided by the dataset curators to create the key-step's list for each task.
The dataset offers dense activity annotations for all the videos.
We created a one-to-many mapping between the key-steps and the provided annotations; this accelerated the annotations process.
The mapping generated was used to create key-step annotations for all videos.
Three people further watched the videos and verified the annotations generated.

To accelerate future research, we release the \dataset dataset on the project web page\textsuperscript{\ref{project_page}}.

\subsection{Task-level details of \dataset}\label{subsec:dataset_stat}
In \Cref{tab:datastats}, we share the statistics for each of the $16$ tasks in the \dataset dataset.
Let $N$ be the number of videos, $K$ be the number of key-steps for a task, $u_n$ be the number of unique key-steps and $g_n$ be the number of annotated key-steps for $n^{th}$ video.
Following~\cite{joint_dynamic_summary}, we calculate the following:

\paragraph{Foreground Ratio} It is the ratio of total duration of the key-steps to the total duration of the video.
This helps to understand the amount of background actions a task has.
The foreground ratio is inversely proportional to the amount of background.
It is calculated as:
\begin{equation}
    F = \frac{\sum_{n=1}^{N}\frac{t_k^n}{t_v^n}}{N}
\end{equation}

Here, $t_k^n$ and $t_v^n$ are the key-step duration and video duration for $n^{th}$ video, respectively.
The range of $F$ is between $0$ and $1$.

From \Cref{tab:datastats}, we can see that the tasks have significant variance in the foreground ratio.
Conversely, tasks like ``PC Assembly'' and ``Tent Assembly'' have a high foreground ratio, suggesting fewer background actions.
On the other hand, tasks like preparing ``Bacon and Eggs'' and ``Turkey Sanwich'' have low foreground ratios, suggesting more background actions.

\begin{table}[!thpb]
    \centering
    \caption{Statistics of the \dataset across different tasks. The high range of the foreground ratio and repeated steps highlights the complexity of the tasks involved in \dataset}
    \newcolumntype{R}{>{\raggedleft\arraybackslash}X}%
    \begin{adjustbox}{width=1\textwidth}
    \begin{tabularx}{\textwidth}{@{}lRRRRR@{}}\toprule
    Task & Videos Count & Key-steps Count & Foreground\newline Ratio & Missing Key-steps & Repeated Key-steps \\
    \midrule
    PC Assembly & $14$ & $9$ & $\mathbf{0.79}$ & $0.02$ & $0.65$\\
    PC Disassembly & $15$ & $9$ & $0.72$ & $\mathbf{0.00}$ & $0.60$\\
    Toy Bike Assembly & $20$ & $\mathbf{17}$ & $0.50$ & $0.06$ & $0.32$\\
    Tent Assembly & $29$ & $12$ & $0.63$ & $0.14$ & $0.73$\\
    Bacon and Eggs & $16$ & $11$ & $0.15$ & $0.22$ & $0.51$\\
    Cheese Burger & $10$ & $10$ & $0.22$ & $0.22$ & $0.65$\\
    Continental Breakfast & $12$ & $10$ & $0.23$ & $0.20$ & $0.36$\\
    Greek Salad & $10$ & $4$ & $0.25$ & $0.18$ & $0.77$\\
    Pasta Salad & $19$ & $8$ & $0.25$ & $0.19$ & $\mathbf{0.86}$\\
    Hot Dog Pizza & $6$ & $8$ & $0.31$ & $0.13$ & $0.62$\\
    Turkey Sandwich & $13$ & $6$ & $0.21$ & $0.01$ & $0.52$\\
    Brownie & $\mathbf{34}$ & $9$ & $0.44$ & $0.19$ & $0.26$\\
    Eggs & $33$ & $8$ & $0.26$ & $0.05$ & $0.26$\\
    Pepperoni Pizza & $33$ & $5$ & $0.53$ & $\mathbf{0.00}$ & $0.26$\\
    Salad & $\mathbf{34}$ & $9$ & $0.32$ & $0.30$ & $0.14$\\
    Sandwich & $31$ & $4$ & $0.25$ & $0.03$ & $0.37$\\
    \bottomrule
    \end{tabularx}
    \end{adjustbox}
    \label{tab:datastats}
\end{table}

\paragraph{Missing Key-steps} This measure captures the count of missed key-steps in each video.
It is defined as:
\begin{equation}
    M = 1 - \frac{\sum_{n=1}^N u_n}{KN}
\end{equation}
The range of $M$ is between $0$ and $1$.
It helps understand if a task can be done even if we miss some steps.
For example, in \Cref{tab:datastats}, ``Salad'' has the highest missing key-steps ratio suggesting that salad can be made if we miss multiple key-steps.
This makes sense, as one can miss adding mayonnaise to the salad but still create an edible salad.
On the other hand, tasks like ``PC Disassembly'' and ``Pepperoni Pizza'' can not afford to miss key-steps as the task won't be complete.
So, for such tasks, we see a missing key-step ratio of $0$.

\paragraph{Repeated Key-steps} This measure captures the repetitions of key-steps across the videos.
It is defined as:
\begin{equation}
    R = 1 - \frac{\sum_{n=1}^N u_n}{\sum_{n=1}^N g_n}
\end{equation}
The range of $R$ is between $0$ and $1$.
Higher values of $R$ indicate repetitions of key-steps across videos.
From \Cref{tab:datastats}, we can see preparing ``Pasta Salad'' has the highest repeated key-steps and preparing ``salad'' has the lowest.
Methods that do not consider repetitions of the key steps, will not perform well for such tasks.
As \meth takes repetitions of the key steps into consideration, it performs well.

\subsection{Datasets not included in \dataset}\label{subsec:excluded}
As mentioned in the main paper, we followed a set of criteria to select videos from existing datasets for including in \dataset. Here we discuss two potential datasets which we could not use for \dataset.

The Charades-Ego dataset~\cite{Sigurdsson2018CharadesEgoAL}, consisting of paired egocentric and third-person videos, is essential for activity recognition. However, it is not practical for procedure learning. The subjects do not perform a series of key-steps to achieve a goal; instead, they perform activities like pouring a drink in a cup and having it. Additionally, the average duration of the videos is $31.2$ seconds compared to $13$ minutes in \dataset, suggesting the briefness of the tasks acted out. 

The EPIC-Kitchens dataset~\cite{Damen2018EPICKITCHENS}, consisting of $100$ hours of kitchen recordings, comes quite close to our requirements. However, due to the unscripted nature of the dataset (which sets it apart from~\cite{egtea_gaze_p}), it becomes unsuitable. As for procedure learning, we need videos of the same tasks performed multiple times.

\section{Applications}\label{sec:applications}

Learning a procedure by observing multiple videos of the same task opens up a range of possible applications.

\mypara{Monitoring procedures} Consider a system trained to know the key-steps for performing a task; if a new person does the same task again, the system will identify if the person misses a step or does a step differently.

\mypara{Guidance systems} A system trained to know the key-steps for performing a task can identify the current step and show the next possible step for performing the task.

\mypara{Automated systems} The proposed framework benefits by enabling automated robotic systems to autonomously learn the key-steps for performing the task by observing the task being performed.
Once the automated system learns the key-steps, the next time, it can do the task without any human assistance.

\section{Additional Experimental Details}\label{sec:experiments_sup}

\subsection{Ablation Results}\label{subsec:egtea_ablation_results}
This section contains ablation results on parts of \dataset.
\Cref{tab:results_ablation_tc3i_sup} contains the results obtained upon replacing the TC3I loss with TCC~\cite{tcc}, LAV~\cite{lav}, and a combination of LAV and TCC~\cite{lav}.
Additionally, \Cref{tab:selecting_k_sup} shows the results obtained upon using various values of $K$.
Finally, \Cref{tab:results_ablation_pcm_diff_loss_cmu} shows the results obtained after considering different combination of losses along with HC and SS for~\cite{CMU_Kitchens,egtea_gaze_p}.

\begin{table}[!tphb]
    \centering
    \caption{\textbf{Effectiveness of the TC3I loss.} 
    Results after replacing TC3I loss in \meth with TCC, LAV, and a combination of LAV and TCC.
    For the majority of the cases, the proposed TC3I loss outperforms all the losses as it focuses on the frame-level correspondences and adds temporal coherency by adopting the C-IDM loss
    }
    \setlength{\tabcolsep}{7.5pt}
    \begin{tabular}{@{}lccc|ccc@{}}\toprule
        \multirow{2}{4em}{Experiment} & \multicolumn{3}{c}{MECCANO~\cite{meccano}} & \multicolumn{3}{|c}{EPIC-Tent~\cite{tent}} \\
        \cmidrule(lr){2-4}\cmidrule(l){5-7}
        & Precision & F-Score & IoU & Precision & F-Score & IoU \\
        \midrule
        TCC+PCM & $15.1$ & $17.9$ & $\mathbf{8.7}$ & $14.2$ & $14.9$ & $7.8$\\
        LAV+TCC+PCM & $13.4$ & $15.6$ & $7.3$ &  $16.0$ & $16.7$ & $\mathbf{8.5}$\\
        LAV+PCM & $14.6$ & $17.4$ & $7.1$ & $15.2$ & $15.8$ & $8.3$\\
        TC3I+PCM (\meth) & $\mathbf{15.5}$ & $\mathbf{18.1}$ & $7.8$ & $\mathbf{17.1}$ & $\mathbf{17.2}$ & $8.3$ \\\bottomrule
        \toprule
        \multirow{2}{4em}{Experiment} &
        \multicolumn{3}{c}{PC Assembly} & 
        \multicolumn{3}{|c}{PC Disassembly} \\
        \cmidrule(lr){2-4}\cmidrule(l){5-7}
        & Precision & F-Score & IoU & Precision & F-Score & IoU \\
        \midrule
        TCC+PCM & $19.9$ & $21.7$ & $11.6$ & $22.0$ & $23.4$ & $12.2$\\
        LAV+TCC+PCM & $21.6$ & $21.1$ & $10.8$ & $21.0$ & $24.3$ & $12.3$\\
        LAV+PCM & $21.5$ & $22.7$ & $11.7$ & $26.4$ & $26.5$ & $12.9$\\
        TC3I+PCM (\meth) & $\mathbf{25.0}$ & $\mathbf{25.1}$ & $\mathbf{12.8}$ & $\mathbf{28.4}$ & $\mathbf{27.0}$ & $\mathbf{14.8}$\\
        \bottomrule
    \end{tabular}
    \label{tab:results_ablation_tc3i_sup}
\end{table}

Consistent with the results obtained in the main paper, in \Cref{tab:results_ablation_tc3i_sup}, we observe highest results when using the proposed TC3I loss.
This is because TC3I accounts for the loss of temporal coherency by TCC~\cite{tcc} with the help of C-IDM loss~\cite{lav}.
Additionally, the TC3I loss focuses on correspondences at the frame level as compared to global alignment employed by LAV~\cite{lav}.

\begin{table}[!htpb]
    \centering
    \caption{\textbf{Selecting the value of $K$.}
    Numbers in \textbf{bold} are highest in the respective row and \underline{underlined} numbers are highest in the respective column}
    \setlength{\tabcolsep}{7pt}
    \begin{tabular}{@{}lrrrr|rrrr@{}}
        \toprule
        \multirow{2}{5em}{Experiment} & \multicolumn{4}{c}{MECCANO~\cite{meccano}} & \multicolumn{4}{|c}{EPIC-Tents~\cite{tent}} \\
        \cmidrule(lr){2-5}\cmidrule(l){6-9}
        &   $K$=$7$ & $K$=$10$ & $K$=$12$ & $K$=$15$ &   $K$=$7$   &  $K$=$10$   &  $K$=$12$   &  $K$=$15$ \\ \midrule
        Random & $\mathbf{13.4}$ & $10.1$ & $8.8$ & $7.4$ & $\mathbf{14.1}$ & $10.6$ & $9.1$ & $8.3$ \\
        TC3I+HC & $\mathbf{16.6}$ & $14.0$ & $11.4$ & $10.8$ & $\mathbf{15.4}$ & $\underline{12.1}$ & $10.6$ & $9.9$ \\
        TC3I+SS & $\mathbf{16.3}$ & $12.6$ & $12.2$ & $10.7$ & $\mathbf{15.9}$ & $11.9$ & $10.7$ & $\underline{10.4}$ \\
        \meth & $\underline{\mathbf{18.1}}$ & $\underline{15.2}$ & $\underline{13.5}$ & $\underline{11.9}$ & $\mathbf{\underline{17.2}}$ & $11.1$ & $\underline{12.1}$ & $9.46$ \\
        \bottomrule
        \toprule
        \multirow{2}{5em}{Experiment} &
        \multicolumn{4}{c}{PC Assembly} & 
        \multicolumn{4}{|c}{PC Disassembly}\\
        \cmidrule(lr){2-5}\cmidrule(l){6-9}
       & $K$=$7$   &  $K$=$10$   &  $K$=$12$   &  $K$=$15$ & $K$=$7$   &  $K$=$10$   &  $K$=$12$   &  $K$=$15$ \\
       \midrule
        Random & $\mathbf{15.1}$ & $11.0$ & $10.4$ & $9.2$ & $\mathbf{15.3}$ & $11.8$ & $10.7$ & $9.6$ \\
        TC3I+HC & $\mathbf{21.7}$ & $17.3$ & $\underline{20.7}$ & $19.2$ & $\mathbf{24.9}$ & $18.3$ & $18.0$ & $20.7$ \\
        TC3I+SS & $\mathbf{24.7}$ & $18.1$ & $18.1$ & $\underline{19.7}$ & $\mathbf{23.6}$ & $19.7$ & $21.0$ & $20.7$ \\
        \meth & $\mathbf{\underline{25.1}}$ & $\underline{18.7}$ & $\underline{20.7}$ & $19.0$ & $\mathbf{\underline{27.0}}$ & $\underline{26.5}$ & $\underline{24.5}$ & $\underline{23.6}$ \\
        \bottomrule
    \end{tabular}
    \label{tab:selecting_k_sup}
\end{table}

\begin{table}[!htbp]
    \centering
    \caption{\textbf{Effectivenss of PCM.} Results after replacing PCM with HC and SS with different losses}
    \setlength{\tabcolsep}{2.8pt}
    \begin{tabular}{@{}lcccc|cccc@{}}\toprule
        \multirow{2}{4em}{Experiment} &
        \multicolumn{4}{c}{CMU-MMAC~\cite{CMU_Kitchens}} & \multicolumn{4}{|c}{EGTEA Gaze+~\cite{egtea_gaze_p}}\\
        \cmidrule(lr){2-5}\cmidrule(l){6-9}
        & Precision & Recall & F-Score & IoU & Precision & Recall & F-Score & IoU \\
        \midrule
        TCC+HC & $17.06$ & $19.47$ & $18.08$ & $8.55$ & $16.78$ & $20.00$ & $18.25$ & $8.33$ \\
        TCC+SS & $17.34$ & $19.71$ & $18.31$ & $8.66$ & $16.96$ & $20.29$ & $18.48$ & $8.18$ \\
        TCC+PCM & $\mathbf{18.46}$ & $\mathbf{21.45}$ & $\mathbf{19.71}$ & $\mathbf{9.46}$ & $\mathbf{17.46}$ & $\mathbf{22.71}$ & $\mathbf{19.74}$ & $\mathbf{8.81}$ \\
        \cmidrule(lr){2-5}\cmidrule(l){6-9}
        LAV+TCC+HC & $17.37$ & $18.40$ & $17.76$ & $8.61$ & $\mathbf{16.59}$ & $19.72$ & $18.02$ & $7.35$ \\
        LAV+TCC+SS & $17.46$ & $17.94$ & $17.57$ & $8.53$ & $16.16$ & $20.05$ & $17.90$ & $7.39$ \\
        LAV+TCC+PCM & $\mathbf{18.80}$ & $\mathbf{21.11}$ & $\mathbf{19.71}$ & $\mathbf{9.03}$ & $16.44$ & $\mathbf{21.40}$ & $\mathbf{18.60}$ & $\mathbf{7.45}$ \\
        \cmidrule(lr){2-5}\cmidrule(l){6-9}
        LAV+HC & $18.44$ & $19.78$ & $19.07$ & $8.66$ & $16.59$ & $18.18$ & $17.35$ & $7.87$ \\
        LAV+SS & $17.82$ & $18.99$ & $18.36$ & $8.53$ & $16.08$ & $18.13$ & $17.04$ & $7.87$ \\
        LAV+PCM & $\mathbf{20.62}$ & $\mathbf{21.95}$ & $\mathbf{21.11}$ & $\mathbf{9.40}$ & $\mathbf{17.42}$ & $\mathbf{21.17}$ & $\mathbf{19.12}$ & $\mathbf{8.02}$ \\
        \cmidrule(lr){2-5}\cmidrule(l){6-9}
        TC3I+HC & $18.47$ & $20.27$ & $19.15$ & $8.98$ & $18.74$ & $23.70$ & $20.82$ & $7.93$ \\
        TC3I+SS & $18.53$ & $21.13$ & $19.66$ & $8.86$ & $17.71$ & $24.09$ & $20.36$ & $7.94$ \\
        \meth & $\mathbf{21.62}$ & $\mathbf{24.38}$ & $\mathbf{22.72}$ & $\mathbf{11.08}$ & $\mathbf{19.58}$ & $\mathbf{24.68}$ & $\mathbf{21.72}$ & $\mathbf{9.51}$ \\
        \bottomrule
    \end{tabular}
    \label{tab:results_ablation_pcm_diff_loss_cmu}
\end{table}

Consistent with our observations in the main paper, in \Cref{tab:selecting_k_sup}, we achieve the highest scores when $K=7$.
Additionally, for most cases, \meth results in the highest scores for all the values of $K$.

\Cref{tab:results_ablation_pcm_diff_loss_cmu} shows the results after using various losses with HC, SS, and PCM for procedure learning~\cite{CMU_Kitchens,egtea_gaze_p}.
Nearly all the experiments using PCM achieve the highest scores for other losses. 
Additionally, we achieve the highest scores with \meth.
Due to the characteristics of TC3I loss and PCM, the results are consistent with our previous observations.

\subsection{Hyper-parameters}\label{subsec:hyper-parameters}
\Cref{tab:hyperparameters} lists the hyper-parametes used for \meth.


\begin{table}[!tphb]
    \centering
    \caption{Hyper-parameter settings for \meth.}
    \setlength{\tabcolsep}{5pt}
    \begin{tabular}{@{}lr@{}}
        \toprule
        Hyper-parameter & Value \\
        \midrule
        No. of key-steps ($K$) & $7$\\
        No. of sampled frames & $32$\\
        Batch Size & $5$\\
        Learning Rate & $10^{-4}$\\
        Weight Decay & $10^{-5}$\\
        Window size ($\sigma$) & $300$\\
        Margin ($\zeta$) & $2.0$\\
        Regularization parameter ($\xi$) & $1.0$\\
        No. of context frames ($c$) & $2$\\
        Context stride & $15$\\
        Embedding Dimension & $128$ \\
        Optimizer & Adam~\cite{adam}\\
        \bottomrule
    \end{tabular}
    \label{tab:hyperparameters}
\end{table}

\bibliographystyle{splncs04}
\bibliography{bibliography}
\end{document}